\documentclass[10pt,twocolumn,letterpaper]{article}

\usepackage{wacv}
\usepackage{times}
\usepackage{epsfig}
\usepackage{graphicx}
\usepackage{amsmath}
\usepackage{amssymb}
\usepackage[justification=centering]{caption}
\usepackage{multirow}
\usepackage{svg}
\DeclareMathOperator{\E}{\mathbb{E}}
\usepackage{graphicx}\graphicspath{{Figures/}{tiff/}}
\usepackage{verbatim}

\wacvfinalcopy 

\def\wacvPaperID{428} 

\ifwacvfinal\fi
\begin{document}

\title{DiscrimNet: Semi-Supervised Action Recognition from Videos using Generative Adversarial Networks}

\author{Unaiza Ahsan \\
Georgia Institute of Technology\\
{\tt\small uahsan3@gatech.edu}
\and
Chen Sun \\
Google\\
{\tt\small chensun@google.com}
\and
Irfan Essa \\
Georgia Institute of Technology \\
{\tt\small irfan@gatech.edu}
}

\maketitle
\ifwacvfinal\thispagestyle{empty}\fi

\begin{abstract}
  We propose an action recognition framework using Generative Adversarial Networks. Our model involves training a deep convolutional generative adversarial network (DCGAN) using a large video activity dataset without label information. Then we use the trained discriminator from the GAN model as an unsupervised pre-training step and fine-tune the trained discriminator model on a labeled dataset to recognize human activities. We determine good network architectural and hyperparameter settings for using the discriminator from DCGAN as a trained model to learn useful representations for action recognition. Our semi-supervised framework using only appearance information achieves superior or comparable performance to the current state-of-the-art semi-supervised action recognition methods on two challenging video activity datasets: UCF101 and HMDB51. 
\end{abstract}


\begin{figure*}[t]
\captionsetup{justification=justified}
\begin{center}
\includegraphics[width=0.9\linewidth,height=5.6cm,keepaspectratio]{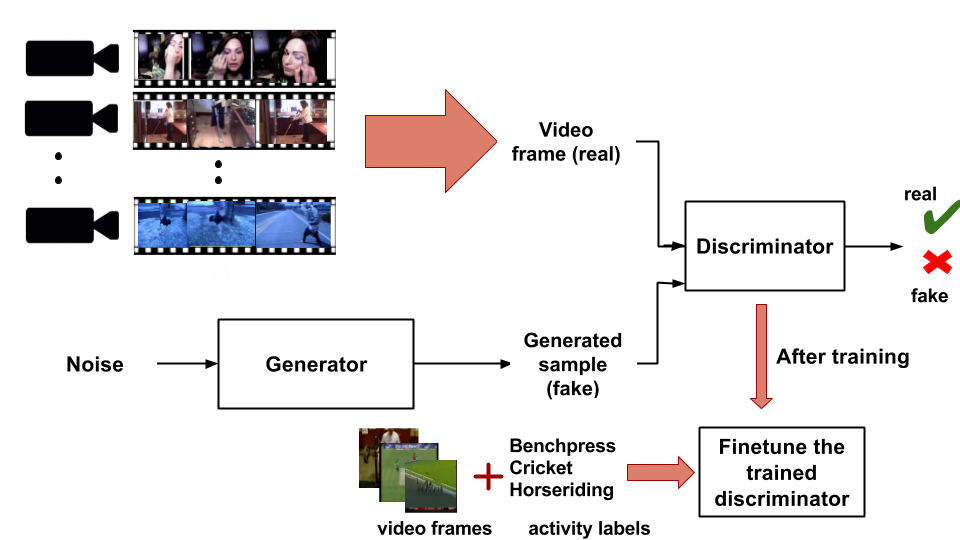}
\end{center}
   \caption{Our approach to learn action representation from GANs}
\label{fig:F1}
\end{figure*}

\section{Introduction}

One of the biggest challenges in recognizing activities in videos is obtaining large labeled video datasets. Annotating videos is largely both expensive and cumbersome due to variations in viewpoint, scale and appearance within a video. This suggests a need for semi-supervised approaches to recognize actions in videos. One such approach is to use deep networks to learn a feature representation of videos without activity labels but with temporal order of frames as a `weak supervision' \cite{misra2016shuffle,fernando2016self}. This approach still requires some supervision in terms of deciding sampling strategies and related video encoding methods to input to neural networks (such as dynamic images \cite{bilen2016dynamic}) and designing `good questions' of correct/incorrect orders as input to the deep network. 

Generative models such as the recently introduced Generative Adversarial Networks (GANs)~\cite{NIPS2014_5423} approximate high dimensional probability distributions like those of natural images using an adversarial process without requiring expensive labeling. To this end, our research question is: \emph{How can we use abundant video data without labels to train a generative model such as a GAN and use it to learn action representation in videos with little to no supervision?}

GANs are conventionally used to learn a data distribution of images starting from random noise. Adversarial learning in GANs involves two networks: a discriminator network and a generator network. The discriminator network is trained on two kinds of inputs -- one consisting of samples drawn from a high dimensional data source such as images and the other consisting of random noise. Its goal is to distinguish between real and generated samples. The generator network uses the output of a discriminator to generate `better' samples. This minimax game aims to converge to a setting where the discriminator is unable to distinguish between real and generated samples. We propose to use the discriminator trained to only differentiate between a real and generated sample for learning a feature representation of actions in videos. 

We use the GAN setup to train a discriminator network and use the learned representation of discriminator as ``initialized weight.'' Then fine-tune that discriminator on labeled video dataset such as UCF101 \cite{soomro2012ucf101}. Recent works have done small experiments \cite{vondrick2016generating} but to our knowledge, nobody has done an in-depth study and especially considered all the architecture/hyperparameter settings that can yield a good performance across datasets (we do well on HMDB51 too) using only appearance information in the video. This unsupervised pre-training step avoids any manual feature engineering, video frame encoding, searching for the best video frame sampling technique and results in an action recognition performance competitive to the state-of-the-art using only appearance information. 

Our key contributions and findings are:

\begin{itemize}
 \item We propose a systematic semi-supervised approach to learn action representations from videos using GANs. 
 \item We perform a comprehensive study of best practices to recognize actions from videos using the GAN training process as a good initialization step for recognition. 
 \item We find that appearance-based unsupervised pre-training for video action recognition performs superior or comparable to the state-of-the-art semi-supervised multi-stream video action recognition approaches.
 \item Our unsupervised pre-training step does not require weak supervision or computationally expensive steps in the form of video frame encoding, video stabilization and search for best sampling strategies. 
\end{itemize}

\section{Related Work}
To date, action recognition is one problem in Computer Vision where deep Convolutional Neural Networks (CNNs) have not outperformed hand-crafted features. Action recognition from videos has come a long way from holistic feature learning such as Motion Energy Image (MEI) and Motion History Image (MHI) \cite{bobick2001recognition}, space-time volumes \cite{yilmaz2005actions} and Action Banks \cite{sadanand2012action} to local feature learning approaches such as space-time interest points~\cite{laptev2005space}, HOG3D~\cite{klaser2008spatio}, histogram of optical flow~\cite{laptev2008learning} and tracking feature trajectories~\cite{messing2009activity,matikainen2009trajectons,jiang2012trajectory,wang2013action}. 

The recent success of CNNs in image recognition has enabled many researchers to treat a video as a set of RGB images, perform image classification on the video frames and aggregate the network predictions to achieve video level classification \cite{simonyan2014two}. Our approach is also inspired by local appearance encoding methods for videos. 3D convolutional networks capture spatio-temporal features via 3D convolutions in both spatial and temporal domains \cite{ji20133d}. Various fusion techniques are proposed to pool the temporal information to construct video descriptors \cite{karpathy2014large, yue2015beyond}. Recurrent Neural Networks (RNNs) and Long Short Term Memory (LSTM) networks have also been used to model videos for action recognition \cite{baccouche2011sequential, donahue2015long}. Using multiple networks to model appearance and motion was first introduced by Simonyan and Zisserman \cite{simonyan2014two}: the two-stream architecture, where the spatial architecture is the standard VGG Net \cite{chatfield2014return} and the temporal stream network takes input stacked optical flow fields. Wu \etal \cite{wu2015fusing} added audio and LSTMs to the network to improve video classification performance. We do not experiment with multiple modalities in this paper as we use only RGB frames as input to the model for our proof of concept. 

Generative models have been successfully used to avoid manual supervision in labeling videos with the most common application being video frame prediction \cite{sutskever2014sequence, zhou2015temporal, vondrick2015anticipating, goroshin2015unsupervised, srivastava2015unsupervised, mathieu2015deep, vondrick2015anticipating, mobahi2009deep, taylor2010convolutional}. Since appearance changes are smooth across videos, temporal consistency \cite{zhang2012slow} and other constraints \cite{jayaraman2015slow} are useful to learn video representations. Our work proposes a generative model as an unsupervised pre-training method for action recognition. While approaches that take temporal coherency into account such as \cite{misra2016shuffle, goroshin2015unsupervised, wang2015unsupervised, wang2015actions} are similar to our work, they are different in that enforcing temporal coherency still involves weak supervision~\cite{misra2016shuffle} where they have to pre-select good samples from a video. We do not do any weak supervision in our approach but only use the generative adversarial training as an unsupervised pre-training step to recognize actions. 

Recently \cite{fernando2016self} train a network to predict the odd video out of a set of videos where the ``odd one out'' is a video with its frames in wrong temporal order. The key difference between our work and theirs is that we do not require any weak supervision in terms of selecting the right video encoding method, sampling strategies or designing effective odd-one-out questions to improve accuracy. Another recent related approach is that of \cite{lee2017unsupervised} where a network is trained to sort a tuple of frames from videos. This sequence sorting task forms the ``unsupervised pretraining'' step and the network is finetuned on labelled datasets. Our approach does not use weak supervision (such as selecting the right tuple via optical flow for example) for the unsupervised pretraining task and uses only appearance information in this work.  

Generative Adversarial Networks \cite{NIPS2014_5423} have been used for semi-supervised feature learning particularly after the introduction of Deep Convolutional GANs (or DCGANs) \cite{radford2015unsupervised}. Radford \cite{radford2015unsupervised} \etal use the discriminator (pre-trained on ImageNet) to compute features on CIFAR10 dataset \cite{krizhevsky2009learning} for classification. Other works to use GANs for semi-supervised learning \cite{chen2016infogan, salimans2016improved, springenberg2015unsupervised, donahue2016adversarial, odena2016semi} are all designed for image recognition, not videos. 

A recent work is \cite{vondrick2016generating} where the authors train GANs for tiny video generation. They fine-tune their trained discriminator model on UCF101 and show promising results. However, their model is significantly more complicated and requires stabilized videos which involves SIFT \cite{lowe2004distinctive} and RANSAC \cite{fischler1981random} computation per video frame, something that is not required by our method which achieves comparable accuracy after finetuning. 

\section{Approach}
We briefly review the main principles behind GAN models and describe our methodology in detail to recognize actions by leveraging their unsupervised feature learning capability on videos.  

\subsection{Generative Adversarial Networks}
GAN networks \cite{NIPS2014_5423} exploit game theoretic approaches to train two different networks; a generator and a discriminator. The generator represented by function $G$ parameterized by $\theta^{(G)}$ starts with an input noise vector $z$ that is sampled from a normal distribution $p_{noise}(z)$, up-samples this noise distribution and outputs an image $\hat{I}$. The discriminator network is a CNN network (represented by function $D$) parameterized by $\theta^{(D)}$ that takes as input an image ($I$ (real image) or $\hat{I}$ (generated or fake image)) and outputs a probability $\in \{0,1\}$ that whether the input image is from the real distribution or generated distribution. Training GANs involve a minimax game in which the generator attempts to `fool' the discriminator into predicting a generated image as real whereas the discriminator attempts to identify correctly which input images are fake. The discriminator cost function is a cross entropy loss defined by:

\begin{equation}
\label{eq1}
\resizebox{1.0 \columnwidth}{!} 
{
$J^{(D)}(\theta^{(D)},\theta^{(G)}) = 
\E_{I \sim p_{data(I)}} [log D(I)] + \E_{z \sim p_{noise(z)}}[\mbox{log}(1-D(G(z)))]$
}
\end{equation}

The minimax objective from Equation~\ref{eq1} can be optimized using gradient-based methods since both discriminator and generator are functions ($D$ and $G$) that are differentiable with respect to their inputs and parameters \cite{goodfellow2016nips}. The solution to this problem is a Nash equilibrium as both functions are trained to minimize their costs while maximizing the other's objective. GANs can be trained using Stochastic Gradient Descent (SGD) with any optimizer of choice. 

\subsection{Training GANs with Video Frames}
So far in the research community, GANs have been primarily used for sample generation. Thus, focus has been on modifying generator parameters, network architecture and loss functions in order to generate higher resolution images with minimal artifacts. The discriminator network in all variants of GANs is trained with binary cross entropy loss (see Equation~\ref{eq1}) \cite{goodfellow2016nips}. Since our focus is not image generation but learning useful features to transfer to the task of action recognition, we are motivated to train and use the discriminator network in GANs for action recognition. The discriminator network in a GAN learns a representation of local appearance features thus modeling objects and scenes in video frames as context. Lastly, it does so in an unsupervised manner i.e. we do not require explicit labels for objects, scenes or actions to pre-train our action recognition model. 

Consider a set of videos ${\mathcal{V}}$ where ${\mathcal{V}} = \left\{V_1,...,V_n\right\}$ and $n$ is the number of videos in the dataset. Each video consists of a variable number of frames (sampled at the rate of one frame per second). We use all the frames in the training set of videos from two challenging video activity datasets without any label information to train the GAN model. Our approach is shown in Figure~\ref{fig:F1}. We train GANs using a variety of techniques proposed in prior research to generate images. To compare with GANs pre-trained on an object recognition dataset, we also train a GAN model on ImageNet \cite{deng2009imagenet} images. We use the same architecture as proposed in the DCGAN \cite{radford2015unsupervised} paper since the authors have demonstrated the transfer learning capability of DCGAN model on CIFAR10 dataset.

\begin{figure}[t]
\captionsetup{justification=justified}
\begin{center}
\includegraphics[width=0.8\linewidth,height=6.0cm,keepaspectratio]{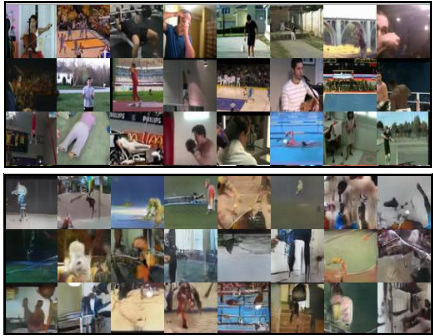}
\end{center}
   \caption{Results after 100 epochs of running DCGAN \cite{radford2015unsupervised} on UCF101 video frames. The images in the top three rows are real while those on the bottom are generated by the model}
\label{fig:F2}
\end{figure}

\subsection{Unsupervised Pre-training}
When dealing with small datasets, a CNN's generalization performance decreases so that the test accuracy remains small even while training accuracy may increase. This is why a common practice is to initialize the weights of the layers with ImageNet pre-trained CNN weights instead of training from scratch. This is referred to as supervised pre-training since ImageNet labels have been used to determine the initial weights. 

Our approach is different in that we are trying to do \textbf{unsupervised} pre-training - determining starting weights for a CNN model (discriminator) which is pre-trained without label information using adversarial training. This unsupervised pre-training setup is compared with initializing the weights in the discriminator network using other settings and we show that the GAN-based initialization significantly outperforms other initialization strategies on the test set of UCF101. 

\subsection{Fine-tuning Discriminator Model}
In this step of our approach we initialize the network with the learned weights from adversarial training and fine-tune it on two video activity datasets. In the process of fine-tuning, we are faced with numerous choices of network architecture, learning rate schemes, optimization and data augmentation. We explore in the space of these variations and report all results on the test split $1$ of UCF101 dataset. 

\begin{figure}[b]
\captionsetup{justification=justified}
\begin{center}
\includegraphics[width=0.9\linewidth,height=3.6cm,keepaspectratio]{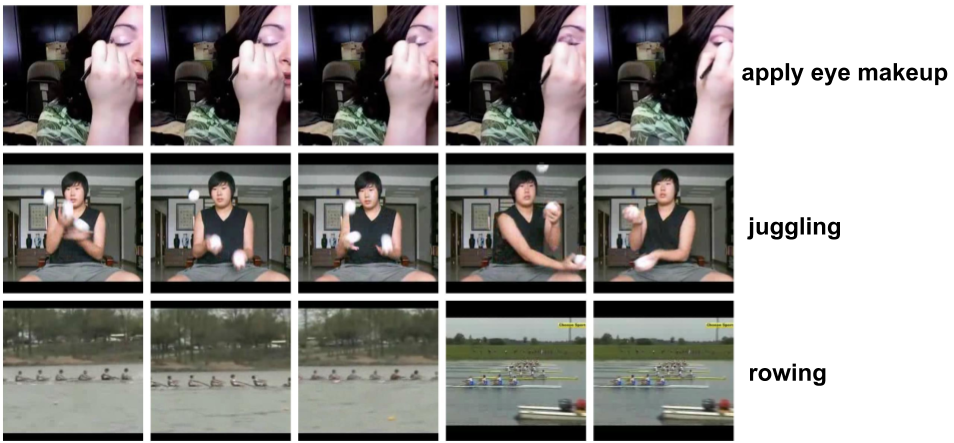}
\end{center}
   \caption{Sample frames from the UCF101 dataset \cite{soomro2012ucf101} with action classes (from top to bottom): apply eye makeup, juggling balls and rowing}
\label{fig:F4}
\end{figure}

\section{Experiments}

\subsection{Datasets}
\begin{figure}[t]
\captionsetup{justification=justified}
\begin{center}
\includegraphics[width=0.9\linewidth,height=3.6cm,keepaspectratio]{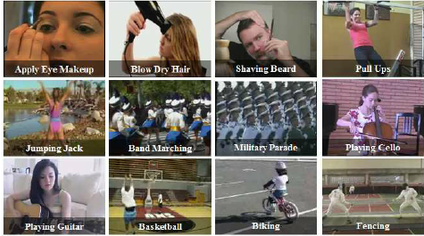}
\end{center}
   \caption{Sample frames from the HMDB51 dataset \cite{Kuehne11}}
\label{fig:F5}
\end{figure}

UCF101 \cite{soomro2012ucf101} is a benchmark action recgonition dataset comprising 13320 YouTube videos of 101 action categories. Actions include human-object interactions such as `apply lipstick', body motion such as `handstand walking', human-human interactions, playing musical instruments and sports. The dataset is small but challenging in that the videos vary in viewpoint changes, illumination, camera motion and blur. The second dataset we experiment on is the HMDB51 dataset \cite{Kuehne11} which contains 6766 videos of 51 actions such as chew, eat, laugh \etc. Sample frames from both datasets are shown in Figures~\ref{fig:F4} and ~\ref{fig:F5}. 

\subsection{Unsupervised Pre-training}
This section describes three experiments to determine: (a) Whether GANs can generate action images (b) Training Protocol of GANs and (c) Data Augmentation steps
\vspace{-3.0mm}

\paragraph{Can GANs Generate Action Images?}
Since we consider a video as a set of RGB frames, we address the first question: Are GANs, traditionally used for generating faces, objects and scenes capable of generating an image representing an action? This question is crucial to address because it determines the validity of using the trained GAN discriminator as a CNN network and fine-tune it on a labelled video activity dataset. To answer this question, we use all the videos from the train split 1 of UCF101 \cite{soomro2012ucf101} and sample 1 frame per second from each video. We train a DCGAN model with default parameters and after 100 epochs, obtain results shown in Figure~\ref{fig:F2}. From visual inspection we can see that vanilla DCGAN is able to learn a coarse representation of activities involving humans. The question now remains whether we can use the feature representation learned by GAN's discriminator as an unsupervised pre-training step to classify actions in labeled video action recognition datasets. 
\vspace{-3.5mm}

\paragraph{Training Protocol of GANs:}
We use DCGAN's public implementation in torch and train three separate GAN models: One with UCF101 video frames, second with ImageNet \cite{deng2009imagenet} images and third with a subset of Sports1M dataset \cite{KarpathyCVPR14} frames. We train all three models for 100 epochs using the architectural guidelines proposed in \cite{radford2015unsupervised}, namely, batch normalization \cite{ioffe2015batch} in discriminator as well as the generator, leaky Rectified Linear Units (leaky ReLU) \cite{xu2015empirical} in all layers of discriminator, strided convolutions in discriminator instead of pooling layers and fractional-strided convolutions in the generator. There are no fully-connected (FC) layers in the DCGAN architecture as the authors of \cite{radford2015unsupervised} report no loss in generator performance for not including FC layers. Hence we also use the same architecture for training the GAN model.
\vspace{-3.0mm}
\paragraph{Data Augmentation:}
The main difference between our GAN training and the DCGAN \cite{radford2015unsupervised} approach is that DCGAN \cite{radford2015unsupervised} performs data augmentation via taking 64 x 64 sized random crops of the image as well as scaling the images to range [-1,1]. This scaling is done for the tanh activation function in the generator. We change that protocol and avoid random cropping. We only scale the frames of videos to the range [-1,1] and scale the size to 64 x 64. The reason why we avoid random cropping is because the action frames from videos are much larger and contain much more information than the original images used for training DCGAN (bedrooms, faces and the like). Taking random crops from action frames will not result in a useful representation because too much information will be lost. Thus, we only scale the images to 64x64 as our aim is not just to generate action images but to learn an effective action representation for recognition. 

\subsection{Fine-tuning for Action Recognition}
\begin{figure*}[t]
\captionsetup{justification=justified}
\begin{center}
\includegraphics[width=1.0\linewidth,height=6.6cm,keepaspectratio]{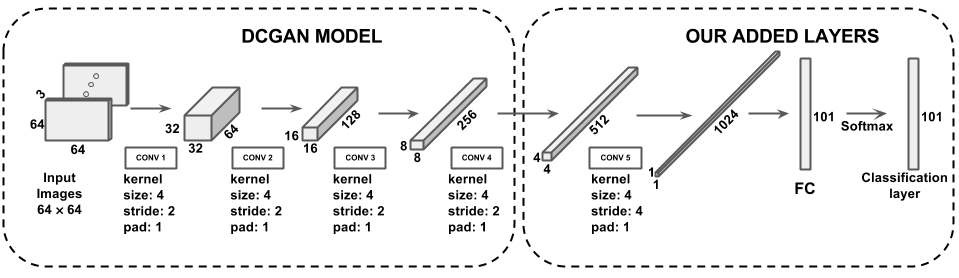}
\end{center}
   \caption{Our network architecture: DCGAN discriminator architecture on the left and our added layers on the right}
\label{fig:F8}
\end{figure*}

Here we describe the set of experiments conducted after the GAN model has been trained. We use the pre-trained discriminator network from our GAN model and fine-tune it on the two labeled video action datasets: UCF101 \cite{soomro2012ucf101} and HMDB51 \cite{Kuehne11}. We begin by replacing the last spatial convolutional layer (CONV5) with one that has the correct number of outputs (equal to the number of action classes). See Figure~\ref{fig:F8}. This layer is initialized randomly and the network is trained again with the previous layers initialized with the pre-trained discriminator's weights.

We perform a comprehensive experimental analysis of architectural choices, hyperparameter settings and other good practices and report the accuracy on the test set of UCF101 dataset.  

\vspace{-3.0mm}

\paragraph{Does Source Data Distribution Matter?}
In this experiment, we determine whether the dataset we train GAN with (which we refer to as the \textit{source dataset}) determines performance on the \textit{target dataset} (the labeled dataset on which we fine-tune the discriminator network). To this end, we train DCGAN on three large scale datasets: ImageNet \cite{deng2009imagenet} images, UCF101 \cite{soomro2012ucf101} video frames and frames of 10,000 videos from Sports1M \cite{karpathy2014large} dataset. We use the same sampling strategy of 1 frame per second for both video datasets and train all three GAN models separately for 100 epochs. 

For each video $V_i$, there is a set of frames $F_i$ where $F_i = {\mathcal{V}} = \left\{f_{n_1},f_{n_2},...,f_{n_i}\right\}$ where $n_i$ is the number of frames extracted for video $V_i$. Each video's frames are passed through the trained GAN's discriminator and we extract CONV4's activations as features on each frame. We average frame-level features to obtain video-level features. We train a linear SVM classifier \cite{fan2008liblinear} on top of these features using the train/test split1 provided by the dataset authors and obtain classification accuracy on the test set. We use the same setting for training all three GAN models as described in the training protocol earlier. Our results are shown in Table~\ref{T1}. 
\begin{table}[t]
\centering
\begin{tabular}{|l|c|c|}
\hline
\multirow{3}{*}{\textbf{Source Dataset}} & \multicolumn{2}{l|}{\textbf{Destination Dataset (accuracy \%)}} \\ \cline{2-3} 
 & \textbf{UCF101} & \textbf{HMDB51}  \\ \hline
ImageNet & 43.88 & 12.82 \\ \hline
UCF101 & \textbf{47.20} & \textbf{12.94} \\ \hline
Sports1M & 42.50 & 13.02 \\ \hline
\end{tabular}
\caption{Comparing the accuracy on target dataset with three large scale datasets used to train GAN models}
\label{T1}
\end{table}

As can be seen from Table~\ref{T1} training a GAN with UCF101 frames results in the best test accuracy on both UCF101 and HMDB51. The difference between training a GAN model with ImageNet and Sports1M frames and training it with UCF101 frames is significant. Note that we did not use all videos from the Sports1M dataset; we randomly selected 10,000 videos from the dataset, extracted 1 frame per second from each video and used those frames to train the GAN model. For HMDB51 dataset the difference in test accuracy between using a GAN discriminator pre-trained on UCF101 and other datasets is not very large. But the superior performance of training a GAN model with video action frames is clearly demonstrated by this experiment. The features learned by the discriminator network are strong enough to transfer to other video datasets as well. 
\vspace{-3.0mm}

\paragraph{Choice of Architecture:}
There are several ways of changing the architecture of the pre-trained discriminator network for fine-tuning. Note that the discriminator is just another CNN network with spatial convolutional layers and no fully connected layers. For fine-tuning on the UCF101 dataset, we replace the last convolutional layer (CONV4) with one that has the correct number of outputs, initialize this layer randomly and train this network (fine-tune) for 160 epochs. This fine-tuning experiment is called `CONV4' in Table ~\ref{T2}. Network depth determines the model's performance both in theory and practice \cite{huang2016deep}. Hence we add another convolutional layer (CONV5) and a fully connected layer (FC), initialize them from scratch and retrain the network till convergence. We extract CONV4, CONV5 and FC features from the finetuned network. We concatenate CONV5 and CONV5 features and test the performance as well as CONV4, CONV5 and FC features. We do not freeze any layers before fine-tuning and keep a learning rate of 0.001 to fine-tune the network. We empirically found that freezing the earlier layers and finetuning only the last layer(s) did not increase performance. We train a linear SVM on top of the extracted features and compute results on UCF101's test set. Our results are shown in Table~\ref{T2}. Our network architecture is shown in Figure~\ref{fig:F8}. 

\begin{table}[b]
\centering
\begin{tabular}{|l|c|}
\hline
\textbf{Architectural changes} & \textbf{Test Accuracy} (\%) \\ \hline
CONV4 & 48.35 \\ \hline
CONV4 + CONV5 + FC & 49.30 \\ \hline
CONV4 + CONV5 & 50.12 \\ \hline
\end{tabular}
\caption{Effect of making the network deeper: Adding more layers slightly improves action recognition performance}
\label{T2}
\end{table}

Our results in Table~\ref{T2} show that with all other parameters kept the same, adding a convolutional layer and a fully connected layer in the discriminator network architecture results in only a slight improvement in performance. We note that this is not a huge difference and this may seem counterintuitive but the reason why this happens is that we are initializing the added network layers randomly before fine-tuning. Also, the dataset size of UCF101 frames is not very large with 84,747 frames in the training set and 33,187 frames in the test set. This may lead to over fitting resulting in only a slight increase in performance on the test set especially when the fully connected layer is added. 

To reduce overfitting, we add dropout \cite{srivastava2014dropout} after the additional convolutional and fully connected layers. We note the performance with/without dropout by extracting CONV4 features from both networks (after finetuning) and training a linear SVM. Adding dropout regularizes the network more thus increasing the performance on test set of UCF101. 

\paragraph{Fine-tuning vs Linear SVM:}
Once we fine-tune the discriminator model on the datasets, we have a choice of whether to extract the CONV4's activations and train a linear SVM on top of it or fine-tune the last layers with softmax classifier. We do both in our experiments and note that the outcome is dependent on the dataset. We find that when we fine-tune the discriminator network on UCF101, the test set accuracy using softmax is lower than extracting CONV4 features and training a linear SVM to recognize actions. However when using HMDB51, the softmax classification on the test set results in a higher accuracy than extracting Layer 9 features and training a linear SVM classifier. This result is shown in Table~\ref{T5}.

\begin{table}[h]
\centering
{%
\begin{tabular}{|l|c|c|}
\hline
\multirow{2}{*}{} & \multicolumn{2}{c|}{Accuracy (\%) on test set} \\ \cline{2-3} 
 & CONV4 + linear SVM & Softmax \\ \hline
UCF101 & \textbf{48.35} & 41.40 \\ \hline
HMDB51 & 14.40 & \textbf{21.04} \\ \hline
\end{tabular}%
}
\caption{Comparing two ways of evaluating fine-tuned network performance on UCF101 and HMDB51 test sets}
\label{T5}
\end{table}

\begin{figure}[t]
\captionsetup{justification=justified}
\begin{center}
\includegraphics[width=0.7\linewidth,height=5.0cm,keepaspectratio]{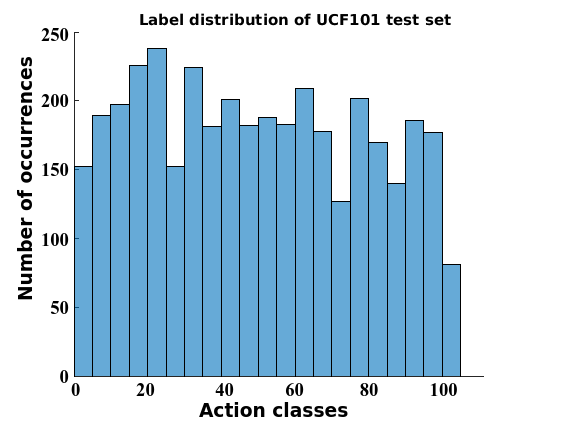}
\end{center}
   \caption{Label distributions of UCF101 test set. The HMDB51 dataset has uniform distribution of 30 videos per action class}
\label{fig:F6}
\end{figure}

From Table~\ref{T5} it is apparent that for UCF101, feature embedding and training a linear SVM results in a better accuracy than softmax classification. The complete opposite is true with HMDB51 dataset. We dig deeper to investigate why this happens. We find that the label distribution of the dataset on which a deep network is being fine-tuned on is the key to determine which method results in a better test accuracy. The label distribution of UCF101 test set is shown in Figure~\ref{fig:F6}. This distribution is not balanced while that of HMDB51 is completely balanced in terms of number of videos per action category. Hence it appears that when classes are unbalanced, since we have not used weighted loss in the neural network, the linear SVM learns the features better hence resulting in an increased performance on the test set. In the case of HMDB51, all classes are balanced equally leading to the superior performance of the softmax classifier over the feature embedding approach.

\vspace{-2mm}

\paragraph{Unsupervised Pretraining vs Random Initialization}
We validate the use of our unsupervised pre-training approach by comparing it with a network that is initialized randomly. We initialize all the layers of the network using `xavier' initialization. Proposed by \cite{glorot2010understanding}, the authors recommend initializing weights by drawing from a distribution with zero mean and variance given by: $Var(W) = 2/(n_{in} + n_{out})$ where $W$ is the distribution which the neuron is initialized with, $n_{in}$ is the number of neurons feeding into the layer and $n_{out}$ is the number of output neurons from this layer. We initialize all layers with this scheme and train the network till convergence on UCF101. For HMDB51, we train a network for 50 epochs with xavier initialized layers and compare that to our proposed discriminator initialized method at 50 epochs. The results are shown in Table~\ref{T8} and clearly validate the use of our unsupervised pretraining approach to initialize the network before finetuning. As a reference, a supervised ImageNet pretrained network finetuned on UCF101 yields an accuracy of 67.1\% and finetuned on HMDB51 yields an accuracy of 28.5\% \cite{misra2016shuffle}.

\begin{table}[]
\centering
\resizebox{\columnwidth}{!}{%
\begin{tabular}{|l|c|c|}
\hline
\textbf{Initialization} & \textbf{UCF101 (\%)} & \textbf{HMDB51 (\%)} \\ \hline
Xavier + finetuning & 33.10 & 11.6 \\ \hline
DiscrimNet (ours) + finetuning & \textbf{49.30} & \textbf{20.4} \\ \hline
\end{tabular}%
}
\caption{Validating the use of our unsupervised pretraining approach vs training with random initialization}
\label{T8}
\end{table}

\subsection{Comparison with the state-of-the-art}
We compare our approach with several recent semi-supervised baselines which recognize actions in videos. The baselines are:
\begin{itemize}
\item \emph{STIP features}: Handcrafted Space Time Interest Point (STIP) features introduced by \cite{laptev2005space}. \vspace{-2.5mm}

\item \emph{DrLim \cite{hadsell2006dimensionality}}: This method uses temporal coherency by minimizing the L2 distance metric between features of neighboring frames in videos and enforcing a margin $\delta$ between far apart frames. \vspace{-2.5mm}

\item \emph{TempCoh \cite{mobahi2009deep}}: Enforce temporal coherencFrom the mid-1980s through 2015 the average number of acres burned has grown from about 2 million acres a year to around 8 millione by using L1 distance instead of L2. Similar to DrLim \cite{hadsell2006dimensionality}. \vspace{-2.5mm}

\item \emph{Obj. Patch \cite{wang2015unsupervised}}: They extract similar object patches using videos and learn a representation of objects by tracking them through time. This model is used and fine-tuned on UCF101 by \cite{misra2016shuffle}. \vspace{-2.5mm}

\item \emph{Shuffle \cite{misra2016shuffle}}: They use sequence verification as an unsupervised pre-training step for vidoes. The model is then fine-tuned on UCF101. \vspace{-2.5mm}

\item \emph{VideoGAN \cite{vondrick2016generating}}: They generate tiny videos using a two stream GAN network. Their model is fine-tuned on UCF101. \vspace{-2.5mm}

\item \emph{O3N \cite{fernando2016self}}: They use odd-one-out networks to predict the wrong temporal order from the right ones. Their model is then fine-tuned on UCF101.\vspace{-2.5mm}

\item \emph{OPN \cite{lee2017unsupervised}}: They train a network to predict the order of 4-tuple frames. Their model is then fine-tuned on UCF101.\vspace{-1.0mm}

\end{itemize}

The results are shown in Table~\ref{T3} and Table~\ref{T7}. 

\begin{table}[t]
\centering
\resizebox{\columnwidth}{!}{
\begin{tabular}{|l|c|}
\hline
\textbf{Method} & \textbf{UCF101-split1 (\%)} \\ \hline
STIP features \cite{laptev2008learning} & 43.9 \\ \hline
DrLim \cite{hadsell2006dimensionality} & 45.7 \\ \hline
TempCoh \cite{mobahi2009deep} & 45.4 \\ \hline
Obj. Patch \cite{wang2015unsupervised} & 40.7 \\ \hline
Shuffle \cite{misra2016shuffle} & 50.9 \\ \hline
VideoGAN \cite{vondrick2016generating} & 52.1 \\ \hline
O3N \cite{fernando2016self} & 60.3 \\ \hline
OPN \cite{lee2017unsupervised} & 56.3 \\ \hline
DiscrimNet (ours) CONV4 + linear SVM & 49.33 \\ \hline
DiscrimNet (ours) CONV5 + linear SVM & 48.88 \\ \hline
DiscrimNet (ours) (CONV4 + CONV5) + linear SVM & 50.12 \\ \hline

\end{tabular}
}
\caption{Comparing our method to state-of-the-art semi-supervised approaches on UCF101}
\label{T3}
\end{table}

\begin{table}[t]
\centering
\begin{tabular}{|l|c|}
\hline
\textbf{Method} & \textbf{HMDB51} (\%) \\ \hline
DrLim \cite{hadsell2006dimensionality} & 16.3 \\ \hline
TempCoh \cite{mobahi2009deep} & 15.9 \\ \hline
Obj. Patch \cite{wang2015unsupervised} & 15.6 \\ \hline
Shuffle \cite{misra2016shuffle} & 19.8 \\ \hline
O3N \cite{fernando2016self} & 32.5 \\ \hline
OPN \cite{lee2017unsupervised} & 22.1 \\ \hline
DiscrimNet (ours) (fine-tuned) & 21.0 \\ \hline
\end{tabular}
\caption{Comparing our method to state-of-the-art semi-supervised approaches on HMDB51}
\label{T7}
\end{table}

\section{Discussion}
Our comparison with several state-of-the-art semi-supervised approaches to recognize actions in vidoes yields important insights. Our results show competitive performance as compared to the state-of-the-art approaches in semi-supervised learning given that:
\begin{itemize}
 \item We only use appearance features and do not experiment with motion content of the video. This is especially intriguing given that our method outperforms STIP features on this dataset. All methods in the results we compare to use temporal coherency as a signal and do motion encoding. \vspace{-2mm}

 \item We do not do weak supervision in the form of temporal consistency and do not design temporal order based networks. The only supervision provided to the GAN is the difference between a real image and noise.\vspace{-2mm}

 \item Our model outperforms several state-of-the-art approaches on HMDB51 given that no video from the dataset was used in the unsupervised pre-training step of this approach. This shows the domain adaptation capability of GAN discriminator networks and that they are able to capture enough information to learn useful representation of actions in video frames. 
\end{itemize}

The methods that outperform our proposed approach are either computationally expensive or require much more supervision in the form of selecting sampling strategies, video encoding methods or in the case of O3N networks \cite{fernando2016self}, designing effective odd-one-out questions for the network to learn feature representations for action recognition. 

\section{Conclusion}
We propose an unsupervised pre-training method using GANs for action recognition in videos. Our method does not require weak supervision in the form of temporal coherency, sampling selection or video encoding methods. Purely on appearance information alone, our method performs either better than or comparable to the state-of-the-art semi-supervised action recognition methods. 

{\small
\bibliographystyle{unsrt}
\bibliography{egbib}
}

\end{document}